\documentclass[journal]{IEEEtran}
\ifCLASSINFOpdf
\else
\fi
\usepackage{amsmath}
\usepackage[multi-part-units=single]{siunitx}
\usepackage{graphicx}
\usepackage{xcolor}
\usepackage{multirow}
\usepackage{bm}
\usepackage{cite}

\begin{document}
\title{Skin Lesion Classification Using CNNs with Patch-Based Attention and Diagnosis-Guided Loss Weighting}

\author{Nils~Gessert,
		Thilo~Sentker,
		Frederic~Madesta,
		R\"udiger~Schmitz,
		Helge~Kniep,
		Ivo~Baltruschat,
		Ren\'e~Werner,
        and~Alexander~Schlaefer% <-this % stops a space   
\thanks{Nils Gessert, and Alexander Schlaefer are with the Institute
of Medical Technology, Hamburg University of Technology, 21073, Hamburg, Germany (e-mail: \{nils.gessert,schlaefer\}@tuhh.de).}
\thanks{Thilo Sentker, Frederic Madesta and Ren\'e Werner are with the Institute of Computational Neuroscience, University Medical Center Hamburg-Eppendorf, 20246 Hamburg, Germany (e-mail: \{t.sentker,f.madesta,r.werner\}@uke.de).}
\thanks{R\"udiger Schmitz is with the Department of Interdisciplinary Endoscopy and Institute of Anatomy and Experimental Morphology, University Medical Center Hamburg-Eppendorf, 20246 Hamburg, Germany (e-mail: r.schmitz@uke.de).}
\thanks{Helge Kniep is with the Department of Diagnostic and Interventional Neuroradiology, University Medical Center Hamburg- Eppendorf, 20246 Hamburg, Germany (e-mail: h.kniep@uke.de).}
\thanks{Ivo Baltruschat is with the Section for Biomedical Imaging, University Medical Center Hamburg Eppendorf, 20246 Hamburg and the Institute for Biomedical Imaging, Hamburg University of Technology, 21073 Hamburg, Germany (e-mail: ivo-matteo.baltruschat@tuhh.de).}
\thanks{All authors are with DAISYlabs, Forschungszentrum Medizintechnik Hamburg, Germany}
\thanks{Copyright (c) 2017 IEEE. Personal use of this material is permitted.
However, permission to use this material for any other purposes must
be obtained from the IEEE by sending an email to
pubs-permissions@ieee.org.}
% <-this % stops a space
}% <-this % stops a space space% <-this % stops a space space


\markboth{IEEE Transactions on Biomedical Engineering, Preprint}%
{Gessert \MakeLowercase{\textit{et al.}}: Skin Lesion Classification Using CNNs with Patch-Based Attention and Diagnosis-Guided Loss Weighting}



\maketitle

\begin{abstract}

\textit{Objective:} This work addresses two key problems of skin lesion classification. The first problem is the effective use of high-resolution images with pretrained standard architectures for image classification. The second problem is the high class imbalance encountered in real-world multi-class datasets.
\textit{Methods:} To use high-resolution images, we propose a novel patch-based attention architecture that provides global context between small, high-resolution patches. We modify three pretrained architectures and study the performance of patch-based attention. To counter class imbalance problems, we compare oversampling, balanced batch sampling, and class-specific loss weighting. Additionally, we propose a novel diagnosis-guided loss weighting method which takes the method used for ground-truth annotation into account.
\textit{Results:} Our patch-based attention mechanism outperforms previous methods and improves the mean sensitivity by $\SI{7}{\percent}$. Class balancing significantly improves the mean sensitivity and we show that our diagnosis-guided loss weighting method improves the mean sensitivity by $\SI{3}{\percent}$ over normal loss balancing.
\textit{Conclusion:} The novel patch-based attention mechanism can be integrated into pretrained architectures and provides global context between local patches while outperforming other patch-based methods. Hence, pretrained architectures can be readily used with high-resolution images without downsampling. The new diagnosis-guided loss weighting method outperforms other methods and allows for effective training when facing class imbalance.
\textit{Significance:} The proposed methods improve automatic skin lesion classification. They can be extended to other clinical applications where high-resolution image data and class imbalance are relevant.

\end{abstract}


% \fi
\IEEEpeerreviewmaketitle

\section{Introduction}

Skin cancer is one of the most frequent types of cancer. With 5 million cases reported annually in the United States it is the most widespread type of cancer \cite{siegel2017cancer,apalla2017skin}. For skin lesion assessment by experts, dermoscopy has been established as an imaging modality that improves diagnostic performance compared to unaided visual examination \cite{kittler2002diagnostic,vestergaard2008dermoscopy}. Typically, experts rely on subjective evaluation of skin lesion features. For more systematic diagnosis procedures, rule sets, such as the "7-point checklist" \cite{argenziano1998epiluminescence} and "ABCD rule" \cite{nachbar1994abcd} have been proposed to reduce interobserver variability.

Furthermore, computer-aided diagnosis (CAD) systems have been proposed that also promise to reduce interoberserver variability and address the limited availability of trained experts \cite{kimball2008us}. Earlier approaches relied on extraction of handcrafted features to be fed into conventional classifiers \cite{maragoudakis2010skin,madooei2012intrinsic}. More recently, deep learning-based approaches have shown tremendous success in the medical imaging domain \cite{litjens2017survey}. A straightforward extension to classic feature extraction is to use deep learning for feature extraction combined with conventional machine learning methods for skin lesion classification\cite{codella2015deep,kawahara2016deep}. More recent approaches moved to end-to-end trainable convolutional neural networks (CNNs) for lesion diagnosis \cite{kawahara2016multi,lopez2017skin,yang2018clinical}. In addition, multi-modal approaches using clinical images, dermoscopy and meta data have been proposed \cite{kawahara20187}, as well as a method where segmentation and lesion structure information is incorporated into the system \cite{diaz2018dermaknet}. Also, 
the work of Esteva et al. \cite{esteva2017dermatologist} represents a cornerstone of skin lesion diagnosis as dermatologist-level performance was achieved by a CNN. In their work, the authors trained the InceptionV3 architecture \cite{Szegedy.2016b} on $\num{130000}$ clinical images and compared its predictions to the assessment of 21 trained dermatologists. While this is a remarkable achievement, the high performance was largely achieved by enormous dataset size with a standard model instead of specific model design for skin lesion diagnosis.

Previous work on skin lesion diagnosis has been evaluated on different datasets, such as different parts of the ISIC database \cite{isicdatabase} which was used for the ISIC 2017: Skin Lesion Analysis Towards Melanoma Detection Challenge \cite{codella2018skin} or the  Argenziano et al. \cite{Argenziano} dataset. As this limits comparability and reproducibility, the HAM10000 (HAM) dataset has been made publicly available \cite{tschandl2018}. The dataset consists of $\num{10015}$ dermoscopic images that can serve as a benchmark for skin lesion diagnosis. HAM was used as a training set for the ISIC 2018 Skin Lesion Diagnosis Challenge where we presented the best approach based on publicly available data \cite{gessert2018skin}. While previous work mostly focused on the binary decision "biopsy" or "no biopsy", the HAM dataset is well suited for a multi-class lesion classification which comes with new challenges to consider.   

The first key problem we identify is the use of high-resolution image information. Typically, images are downsampled to a lower input resolution for CNNs, as memory and computational resources are limited. The downsampling process implies that fine-grained image information is lost which might be crucial in a medical context. Previously, multi-resolution approaches addressed this problem for segmentation where the region of interest (ROI) is known a priori due to pixel-level labels \cite{Kamnitsas.2017}. Here, multi-resolution crops can be created around the ROI. However, for skin lesion diagnosis, the relevant ROI is not necessarily known as image-level labels are used instead of pixel-level labels. Kawahara et al. \cite{kawahara2016multi} used a two-path CNN with two input resolutions of the entire lesion image. Due to the high image resolution, this requires extensive downsampling or pooling inside the CNN's high-resolution path which, again, might remove relevant features. Also, preliminary lesion segmentation has been used to determine the relevant ROI \cite{diaz2018dermaknet}. However, this step might cut our relevant parts due to segmentation errors or, depending on lesion size, the ROI might still be too large for conventional CNNs.  

Furthermore, patch-based approaches that use small crops from the high-resolution images as the input to a CNN can be used as often practiced in the natural image domain. To capture the entire image with small crops, traditionally, multi-crop evaluation is used where the predictions from all crops are combined, e.g., through averaging or voting. This approach can be advantageous as using small patches is computationally cheap and, crucially, pretrained standard architectures from the natural image domain with typically small input sizes ($224\times 224$) can be used. Transfer learning with pretrained models has been proven to substantially increase performance for medical learning problems \cite{hoo2016deep,tajbakhsh2016convolutional}. However, multi-crop evaluation can be challenging as local patches need to be combined in a meaningful way. Simple methods such as averaging and voting treat all patches equally which is problematic for skin lesion classification as the lesion will only cover a part of the image. In the natural image domain, this problem has been approached using recurrent architectures which recover global context between local, spatial features \cite{zhu2017dependency,li2016lstm}. We address this problem with a new \textit{patch-based attention} method that uses an attention mechanism to learn global context between patches in an end-to-end trainable model. Inspired by channel-wise attention \cite{hu2018squeeze}, we augment an architecture by a simple block that uses attention across patches. The block is flexible and can be plugged into any pretrained standard architecture. In part of this method we propose \textit{patch dropout} as an effective regularization method. We show that our method substantially improves performance over naive aggregation with averaging and outperforms previous methods using recurrent models for global context. 

Second, we consider class imbalance as a key property of multi-class skin lesion datasets that needs to be addressed. While sampling approaches to counter class imbalance have been used for melanoma detection \cite{kawahara20187}, there is no systematic analysis of different methods for multi-class skin lesion classification. We consider both balanced sampling and variants of loss weighting based on class frequency. As a part of the analysis, we propose a new \textit{diagnosis-guided loss weighting} strategy that includes the method of diagnosis that was used for ground-truth annotation of a lesion. In case a more costly method such as histopathology was used for diagnosis of a benign lesion, we assume that image classification is difficult and thus, we put a higher weight on that example within the loss function. We show that this new incorporation of meta information into the loss function outperforms previous balancing methods.

In support of reproducible research, we make our code and training/validation/test splits publicly available\footnote{https://github.com/ngessert/patch-lesion}. % 

\section{Methods} \label{sec:methods}

\begin{figure}%[!t]
\centering
\includegraphics[width=0.45\textwidth]{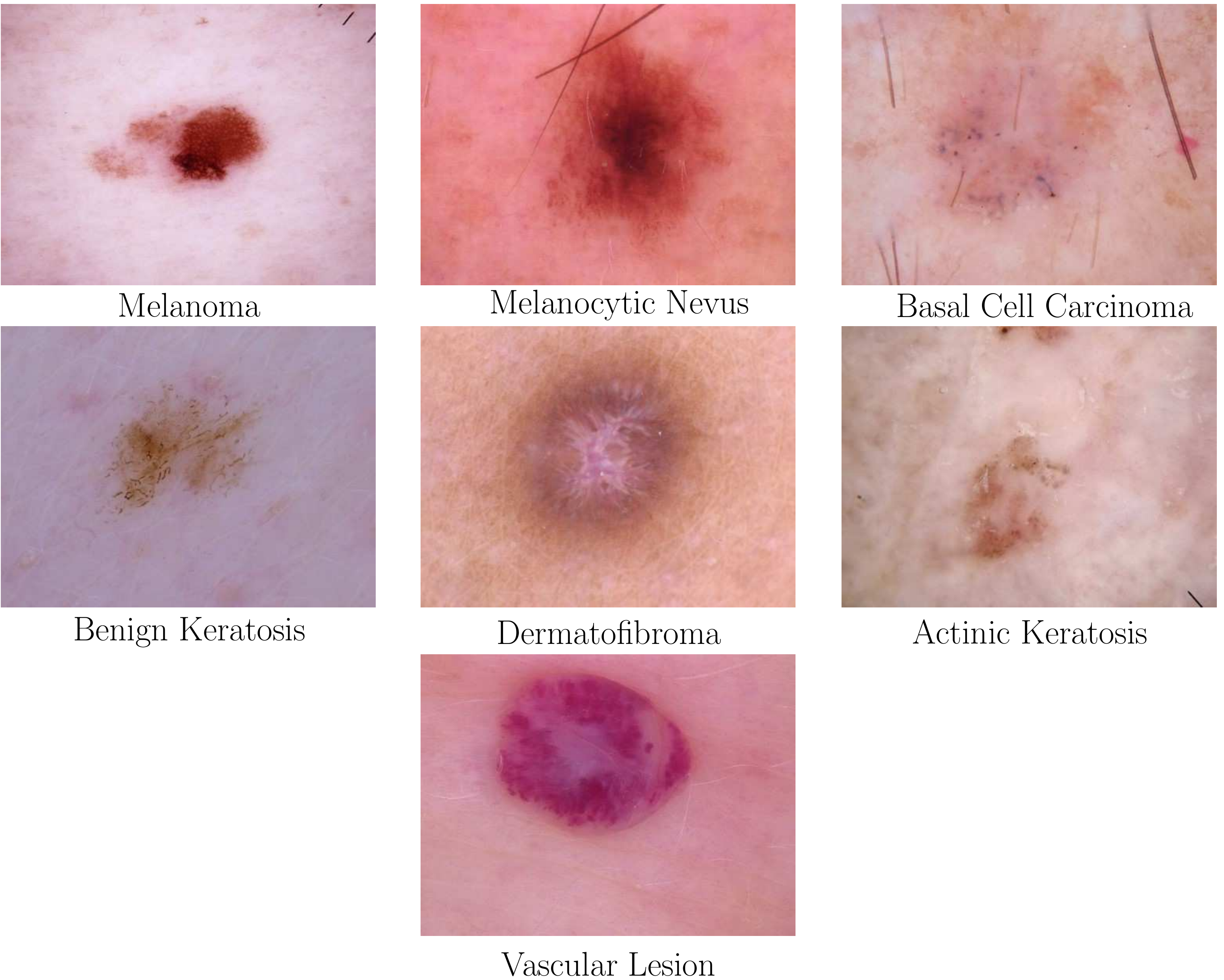}
\caption{Examples for the seven classes in the HAM dataset.}
\label{fig:classes}
\end{figure}

\subsection{Datasets} \label{sec:dataset}

Throughout this work we use the HAM dataset for training and evaluation. The dataset contains images to be classified into seven mutually exclusive classes, see Fig.~\ref{fig:classes}. The dataset distribution represents a real-world setting with overall more benign images while malignant cases are still over-represented. The dataset class distribution is shown in Table~\ref{tab:dist} which highlights the inherent class imbalance as a key problem to be addressed. Also, other public datasets show a similar unbalanced characteristic. 

\begin{table}
	\caption{The size of the three data sets HAM, "7-point criteria" \cite{kawahara20187} (SPC) and the ISIC archive \cite{isicdatabase} (ISIC). All three show a highly imbalanced class distribution. The lesion types are melanoma (MEL), melanocytic nevus (NV), basal cell carcinoma (BCC), actinic keratosis (AKIEC), benign keratosis (BKL), dermatofibroma (DF), and vascular lesions (VASC).}
	\label{tab:dist}
	\centering
	\begin{tabular}{l l l l l l l l}
	 & MEL & NV & BCC & AKIEC & BKL & DF & VASC \\ \hline \\
	 HAM & $1113$ & $6705$ & $514$ & $327$ & $1099$ & $115$ & $142$ \\
	 SPC & $252$ & $575$ & $42$ & $0$ & $69$ & $20$ & $29$ \\
	 ISIC & $1056$ & $11861$ & $72$ & $2$ & $477$ & $7$ & $0$ \\
	\end{tabular}
\end{table}

We split HAM into four equally-sized parts, each exhibiting the same class distribution. One of the parts is held out as a test set ($\num{2500}$ examples). The other three parts are used for three-fold cross-validation which is used for hyperparameter tuning ($\num{5000}$ training examples, $\num{2500}$ validation examples). All results are reported for the independent test set. Note that histogram equalization was used by the dataset publishers for some images \cite{tschandl2018}. Also, the dataset publishers manually centered the images around the lesion and downsampled and resized them to a resolution of $600 \times 450$. We do not employ any additional preprocessing on the HAM images. We do not compare to the associated ISIC 2018 Challenge test set, as corresponding ground-truth data is currently not available to the public. Moreover, the methods presented at the challenge lack comparability due to use of different datasets, differently sized ensembles, varying models and test-time data augmentation.

We demonstrate the generalizability of our method by also applying the proposed attention mechanism to the 7-point criteria (SPC) dataset \cite{kawahara20187} which is based on the dataset presented in \cite{Argenziano}. We match the classes in SPC to the seven HAM classes which leads to some images being dropped if they do not match one of the classes. Also, the dataset does not contain examples for actinic keratosis. Therefore, we train on six classes for experiments with this dataset. We split off a test set with $395$ images, following \cite{kawahara20187}, and train on the remaining $616$ images. We use the hyperparameters we found with cross-validation on HAM and therefore do not use a validation set with this dataset. We resize the images to $600 \times 450$ such that they match HAM in terms of size. In contrast to the HAM preprocessing, we do not center the images around the lesion and we do not apply color correction. 
\subsection{Metrics} \label{sec:metrics}

Regarding evaluation metrics, the nature of the multi-class lesion classification problem has to be taken into account. A normal accuracy would favor and encourage the correct classification of overrepresented classes which is critical, considering the unbalanced dataset. Therefore, we use the multi-class sensitivity (MC-Sensitivity) $S$ for ranking approaches which is defined as

\begin{equation}
	S = \frac{1}{C}\sum_{i=1}^{C}\frac{\mathit{TP}_i}{\mathit{TP}_i+\mathit{FN}_i}
\end{equation}

where $\mathit{TP}$ denotes true positives, $\mathit{FN}$ denotes false negatives and $C$ denotes the number of classes. This metric is challenging in particular, as the dataset is highly imbalanced while the metric treats all classes equally. Besides the MC-sensitivity, we also report MC-specificity and the F1-score. 

\subsection{Baseline Models} \label{sec:baseline}

We consider three architectures for our experiments. 
The first baseline model is Inception-V3 \cite{Szegedy.2016b} (InceptV3) which is a popular choice for medical image analysis \cite{Litjens.2017} and has been applied successfully to skin lesion classification tasks \cite{esteva2017dermatologist,kawahara20187}. This model consists of different Inception modules that process feature maps in parallel with different convolutional paths. 
The second model is the more recent Densenet-121 (Dense121) which belongs to the group of densely connected CNNs \cite{huang2017densely}. The key idea of this architecture is to reuse all previous layer outputs as the input to each layer. Thus, the result of the $l^{th}$ layer is defined by $x_l = H([x_0,x_1,...,x_{l-1}])$ where $x_l$ is a feature map and $H$ is a mapping including convolutions, batch normalization \cite{Ioffe.2015} and the ReLu activation function. To keep feature map growth bounded, the model is subdivided into blocks with transition layers in between that reduce the feature map dimension. 
The third model is the recent SE-Resnext50 (SE-RX) which is based on the multi-path Resnext architecture \cite{Xie.2017} and augmented by attention-like feature recalibration modules \cite{hu2018squeeze}. These \textit{squeeze and excitation} (SE) modules explicitly learn relationships between feature maps which increases representational power over the classic, implicit feature map summation in convolutional layers. 

\subsection{Model Input Strategies} \label{sec:input_strategy}

We consider different strategies to map the images to the standard model input size of $224 \times 224$ \cite{huang2017densely,hu2018squeeze}.

\textbf{Downsampling.} This method represents a simple baseline where the whole image is downsampled to the models' input size. This is used both for training and evaluation.

\textbf{Single-Crop.} Here, images are randomly resized and then cropped to the models' input size during training. Thus, we induce more variation during training. For evaluation, a center crop that covers $\SI{85}{\percent}$ of the image is taken and resized to the model input size. This method is similar to the strategies used for the models SE-RX, Dense121 and InceptV3 \cite{Szegedy.2016b,huang2017densely,hu2018squeeze} and provides a second baseline. 

\textbf{Multi-Crop.} For this method we do not resize the images and randomly crop patches of the models' input size from the image. During evaluation, we use ordered cropping where each patch location is fixed at a predefined point in the image. Then, we average over the predictions of all crops. The number of crops is $N_C \in \{5,9,16\}$ where $5$ covers the four corners and the center and $9$ and $16$ are evenly distributed over the image with overlap between patches. This method has been successfully used for skin lesion classification \cite{gessert2018skin}. 

\textbf{Ordered-Crop.} This strategy uses fixed patch locations during evaluation and training. As patch locations are fixed during training, there is less variation compared to single and random cropping. Therefore, we propose \textit{patch dropout} to avoid overfitting. For each mini-batch update during training, each patch is zeroed out with a probability of $p_d$. We also study the effect of different values for $p_d$. 

\subsection{Local and Global Information Aggregation Strategies}

\begin{figure*}%[!t]
\centering
\includegraphics[width=1.0\textwidth]{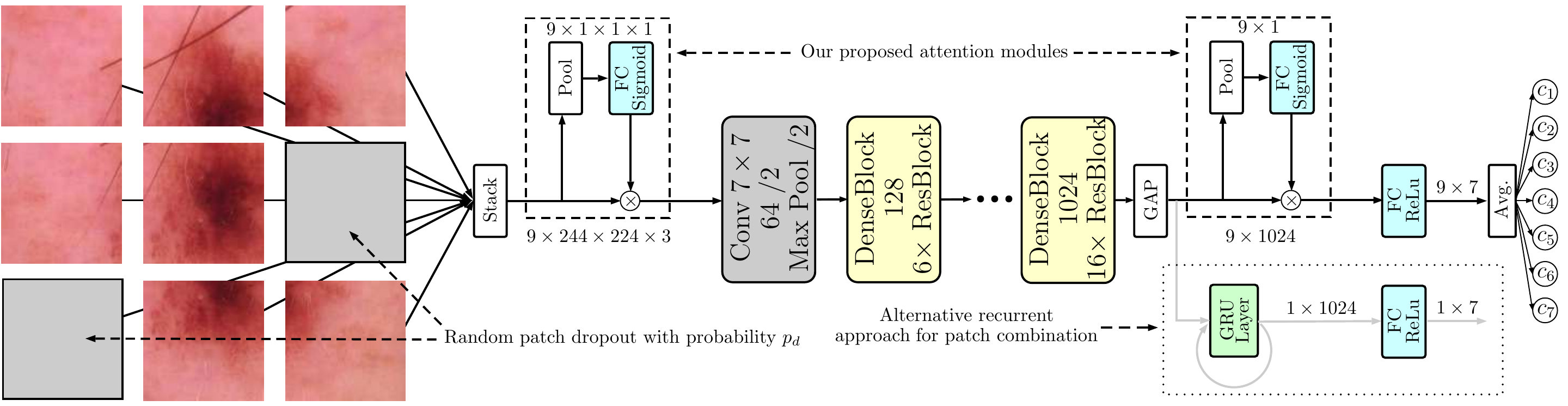}
\caption{The proposed architecture, shown for the example of a Densenet \cite{huang2017densely}. Both our attention approach and recurrent patch combination are shown. Only one of those is applied at a time. For all tensor shapes, the batch size is omitted. Note that the crops are overlapping. Grey patches indicate our novel patch dropout regularization. GAP denotes global average pooling.}
\label{fig:attention}
\end{figure*}

The previously described cropping strategies combine fine-grained, local and global image information in different ways. \textit{Downsampling} directly processes images at a global scale. \textit{Single-Crop} implicitly considers local and global information during training by randomly resizing the images to different scales. The single center crop for evaluation represents a mixture of a local and a global scale. \textit{Multi-Crop} leverages local information during training by cropping small patches from the fully-sized images. This maximizes the extraction of fine-grained, local features. However, global information is only weakly covered by averaging over multiple, equally-weighted patches. The method only aggregates isolated, local context and does not include dependencies between patch locations and the importance of the individual patches. This is particularly problematic for dermoscopy images as patches at the image borders might not cover the lesion at all.

Therefore, we study methods that explicitly combine both local and global image information. We still use small patches and we recover global context between patches with a novel patch-based attention mechanism as shown in Fig.~\ref{fig:attention}.

\begin{figure}%[!t]
\centering
\includegraphics[width=0.49\textwidth]{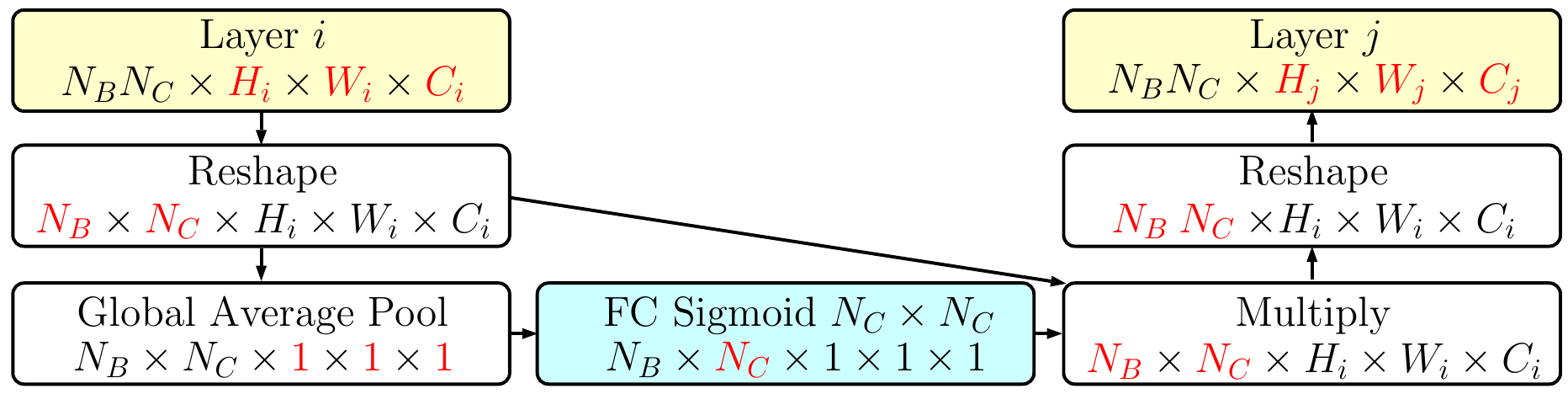}
\caption{Our new attention block in detail. Each block represents an operation. The output tensor size is given in each block. Red indicates the relevant dimensions for each operation.}
\label{fig:attention_details}
\end{figure}

The attention mechanism is illustrated in more detail in Fig.~\ref{fig:attention_details}. The original model input is a tensor of size $N_B \times N_C\times H\times W\times C$ where $N_B$ is the batch size, $N_C$ is the number of crops and $H\times W\times C$ are the crop dimensions. Then, we stack all crops in the batch dimension such that the model input is of size $N_BN_C\times H\times W\times C$. In this way, parameters are shared for patch processing as it is the case for standard multi-crop evaluation. If we reach an attention module after layer $i$, we reshape $N_B$ and $N_C$ back into individual dimensions such that the feature tensor is of size $N_B \times N_C\times H_i\times W_i\times C_i$. Then, the module is applied as follows. We create the initial patch weights $w_i$ of size $N_B\times N_C$ with global average pooling over the last three dimensions of the feature tensor. Next, a fully-connected attention layer $f^{a}_{i}$ with learnable parameters $\theta^{a}_{i}$ of size $N_C \times N_C$ transforms the patch weights. A sigmoid activation function $\sigma$ squashes the transformed weights to $\sigma (f^{a}_{i}(w_i)) \in [0,1]$. Then, the transformed and squashed weights of size $N_B\times N_C$ are multiplied element-wise with the original feature tensor of size $N_B \times N_C\times H_i\times W_i\times C_i$. Thus, the patches are reweighted by learned interaction between each of them. Eventually, the tensor is again reshaped to $N_BN_C\times H_i\times W_i\times C_i$ for simultaneous processing in the network. This process can be repeated at more locations inside the network at any layer $j$. At the output, we average over the crops' predictions.

This processing strategy is computationally efficient as all crops are processed by the same network. Also, we can incorporate patch-based attention modules in any pretrained model at any location since they only operate on the architecture-independent batch dimension of tensors. However, due to different internal model structure, the effect of using different locations will not be comparable between architectures. Thus, in order to ensure comparability between the different model architectures, we examine the effect of introducing attention modules at the beginning and/or at the end of the respective models.

We compare this approach to global context recovery through recurrent models which has been used previously \cite{zhu2017dependency,li2016lstm}. Here, we place a gated recurrent unit (GRU) \cite{cho2014learning} as a replacement for our attention block at the output. The feature vector from each patch is fed into the GRU which learns a combination of these features which is passed to the fully-connected layer. We considered multi-layer GRUs, bidirectional variants and different feature map sizes. Based on cross-validation, we found a single layer with $1024$ feature maps to perform best.

\subsection{Balancing Strategies} \label{sec:balancing}

Next, we address class imbalance in the dataset with four different approaches. 

\textbf{Oversampling.} For this method we repeat the samples of each class in our training set such that all classes have an equal amount of samples to draw from. During training, we sample uniformly and randomly from this new pool of samples. In this way, an equal number of samples from each class will be present throughout the entire training process.

\textbf{Balanced batches.} The oversampling approach does not guarantee an equal number of samples in each batch, only an approximation over the entire training. Therefore, we also consider strictly balanced sampling where each batch is constructed such that it contains exactly the same number of samples from each class.

\textbf{Loss weighting.} Here, we increase the weight of examples from underrepresented classes with a factor of $n_i = \left(N/N_i\right)^k$ where $n_i$ is the weight for class $i$, $N$ is the total number of examples, $N_i$ is the number of examples in class $i$. The exponent $k$ controls the strength of the weighting. We explore different values with $k \in \{0.5,1.5\}$ and our baseline value is $k=1$.

\textbf{Diagnosis-guided loss weighting.} This new balancing strategy incorporates the knowledge of the method used for ground-truth annotation into the loss function. In general, the images in HAM have been labeled based on "expert consensus", "serial imaging showed no change", "confocal microscopy" and "histopathology". We hypothesize that these categories indicate the difficulty of examination, for example, a case of nevus that needed to be evaluated with histopathology is likely hard to identify from the image only. Therefore, we propose to increase the loss of examples based on their ground-truth annotation technique. Specifically, we increase the loss for benign cases when more costly methods have been employed. For example, benign cases that have been evaluated based on expert consensus receive a lower weighting and benign cases evaluated by histopathology receive a higher weighting.

\subsection{Training}

During training, we employ online data augmentation. We randomly flip images along both axes and randomly change brightness and saturation. We use a cross-entropy loss function. Stochastic gradient descent is performed with Adam \cite{Kingma.2014}. We implement our models in PyTorch \cite{paszke2017automatic}. Further implementation details can be found in our publicly available code.

\subsection{Experiments} \label{sec:experiments}

\begin{figure*}[!htb]
	\centering
	\includegraphics[width=1.0\linewidth]{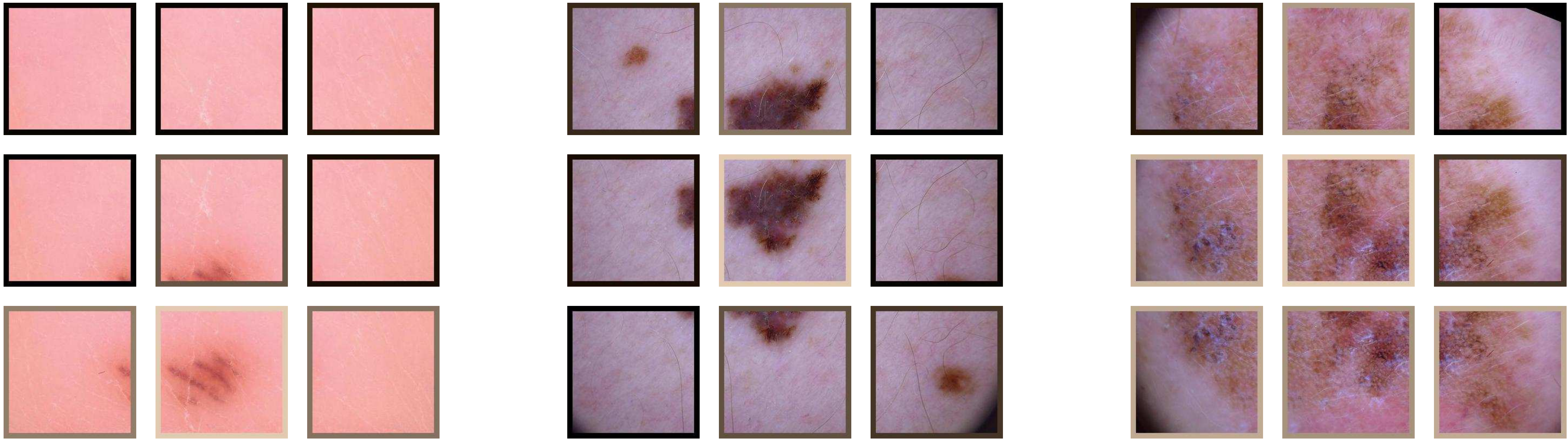}
   \caption{Visualization of the attention weights for three images for $N_C = 9$ and attention at the end of the network. The patch borders represent the relative weighting. Brighter colors represent a higher weighting, darker colors represent a lower weighting.}
\label{fig:vis}	
\end{figure*}

\begin{table}[!t]
\renewcommand{\arraystretch}{1.3}
\caption{Results for the patch-based attention method compared to other methods. Initial Attention uses an attention block at the model input, End Attention uses an attention block at the end of the model, Dual Attention uses both.}
\label{tab:resatt}
\centering
\begin{tabular}{l l l l l}
 & & MC-Sensitivity & MC-Specificity & F1-Score \\
\hline
\parbox[t]{2mm}{\multirow{7}{*}{\rotatebox[origin=c]{90}{InceptV3}}} & Downsampling & $63.3 \pm 2.4$ & $94.6 \pm 0.2$ & $78.7 \pm 0.8$ \\
& Single-Crop & $67.2 \pm 2.1$ & $94.9 \pm 0.2$ & $79.8 \pm 0.8$ \\
& Multi-Crop & $68.4 \pm 2.8$ & $95.5 \pm 0.2$ & $78.2 \pm 0.2$ \\
& GRU & $69.4 \pm 0.3$ & $95.6 \pm 0.3$ & $82.4 \pm 0.7$ \\
& Initial Attention & $72.4 \pm 1.3$ & $96.0 \pm 0.2$ & $84.0 \pm 0.2$ \\
& \textbf{End Attention} & $\bm{73.3 \pm 2.1}$ & $96.0 \pm 0.3$ & $85.3 \pm 1.3$ \\
& Dual Attention & $72.0 \pm 2.0$ & $96.0 \pm 0.2$ & $85.1 \pm 0.5$ \\
\hline
\parbox[t]{2mm}{\multirow{7}{*}{\rotatebox[origin=c]{90}{Dense121}}} & Downsampling & $66.6 \pm 0.7$ & $95.3 \pm 0.0$ & $81.2 \pm 0.2$ \\
& Single-Crop & $70.1 \pm 0.9$ & $95.8 \pm 0.1$ & $81.9 \pm 0.3$ \\
& Multi-Crop & $71.1 \pm 1.1$ & $96.1 \pm 0.1$ & $82.1 \pm 0.3$ \\
& GRU & $72.4 \pm 1.6$ & $95.9 \pm 0.1$ & $82.5 \pm 0.1$ \\
& Initial Attention & $74.8 \pm 0.8$ & $96.1 \pm 0.2$ & $81.7 \pm 0.5$ \\
& End Attention & $74.0 \pm 0.7$ & $96.0 \pm 0.1$ & $82.5 \pm 1.0$ \\
& \textbf{Dual Attention} & $\bm{75.7 \pm 1.3}$ & $96.0 \pm 0.0$ & $82.6 \pm 1.3$ \\
\hline
\parbox[t]{2mm}{\multirow{7}{*}{\rotatebox[origin=c]{90}{SE-RX}}} & Downsampling & $65.0 \pm 1.2$ & $95.1 \pm 0.1$ & $79.5 \pm 0.6$ \\
& Single-Crop   & $69.8 \pm 1.7$ & $95.4 \pm 0.1$ & $77.1 \pm 0.2$ \\
& Multi-Crop & $70.2 \pm 1.0$ & $96.0 \pm 0.0$ & $81.2 \pm 0.7$ \\
& GRU & $71.6 \pm 1.6$ & $95.9 \pm 0.2$ & $83.8 \pm 1.5$ \\
& Initial Attention & $73.2 \pm 2.0$ & $95.8 \pm 0.2$ & $82.5 \pm 1.0$ \\
& \textbf{End Attention} & $\bm{73.7 \pm 0.8}$ & $96.1 \pm 0.2$ & $83.4 \pm 0.1$  \\
& Dual Attention & $73.3 \pm 0.5$ & $96.0 \pm 0.0$ & $82.6 \pm 1.3$  \\
\hline
\end{tabular}
\end{table}

First, we study our novel patch-based attention method. We use loss weighting with all models. All results are for HAM unless stated otherwise. We consider the following experiments.
\begin{itemize}
\item We compare different input strategies for local and global information processing.
\item We visualize the learned attention weights.
\item We compare different numbers of crops being used.
\item We analyse patch dropout with different values for $p_d$.
\item We compare performance on the SPC dataset.
\end{itemize}
Second, we address class balancing strategies. Our novel diagnosis-guided loss weighting strategy is used in conjunction with loss weighting. We use the Multi-Crop models with $N_C=9$. We consider the following experiments.
\begin{itemize}
\item We compare different balancing approaches.
\item We analyze the class balancing strength with loss weighting and varying $k$.
\end{itemize}

\section{Results} \label{sec:results}

\begin{table}[!t]
\renewcommand{\arraystretch}{1.3}
\caption{Results for different numbers of crops from the fully-sized image. Attention refers to the model with an attention block at the beginning of the model. The number indicates the amount of crops.}
\label{tab:numcrops}
\centering
\begin{tabular}{l l l l l}
 & & MC-Sensitivity & MC-Specificity & F1-Score \\
\hline
\parbox[t]{2mm}{\multirow{6}{*}{\rotatebox[origin=c]{90}{InceptV3}}} & Multi-Crop 5 & $63.5 \pm 3.2$ & $95.0 \pm 0.2$ & $75.3 \pm 0.6$ \\
& Multi-Crop 9 & $68.4 \pm 2.8$ & $95.5 \pm 0.2$ & $78.2 \pm 0.2$ \\
& Multi-Crop 16 & $70.8 \pm 1.8$ & $95.9 \pm 0.1$ & $80.1 \pm 0.8$ \\
& Attention 5 & $70.2 \pm 0.6$ & $95.4 \pm 0.3$ & $82.8 \pm 0.8$ \\
& \textbf{Attention 9} & $\bm{72.4 \pm 1.3}$ & $96.0 \pm 0.2$ & $84.0 \pm 0.2$ \\
& Attention 16 & $72.3 \pm 1.1$ & $95.8 \pm 0.1$ & $83.4 \pm 0.5$ \\ 
\hline
\parbox[t]{2mm}{\multirow{6}{*}{\rotatebox[origin=c]{90}{Dense121}}} & Multi-Crop 5 & $67.4 \pm 0.4$ & $95.7 \pm 0.0$ & $79.8 \pm 0.7$ \\
& Multi-Crop 9 & $71.1 \pm 1.1$ & $96.1 \pm 0.1$ & $82.1 \pm 0.3$ \\
& Multi-Crop 16 & $73.1 \pm 1.6$ & $96.3 \pm 0.2$ & $83.7 \pm 1.1$ \\
& Attention 5 & $72.7 \pm 1.4$ & $95.8 \pm 0.2$ & $80.9 \pm 0.7$ \\
& Attention 9 & $74.8 \pm 0.8$ & $96.1 \pm 0.2$ & $81.7 \pm 0.5$ \\
& \textbf{Attention 16} & $\bm{75.2 \pm 0.5}$ & $96.0 \pm 0.0$ & $78.0 \pm 0.4$ \\
\hline
\parbox[t]{2mm}{\multirow{6}{*}{\rotatebox[origin=c]{90}{SE-RX}}} & Multi-Crop 5 & $66.4 \pm 1.4$ & $95.4 \pm 0.1$ & $78.5 \pm 1.1$ \\
& Multi-Crop 9 & $70.2 \pm 1.0$ & $96.0 \pm 0.0$ & $81.2 \pm 0.7$ \\
& Multi-Crop 16 & $72.2 \pm 1.6$ & $96.3 \pm 0.1$ & $82.2 \pm 0.5$ \\
& Attention 5 & $72.0 \pm 1.8$ & $95.7 \pm 0.1$ & $82.9 \pm 0.6$ \\
& Attention 9 & $73.2 \pm 2.0$ & $95.8 \pm 0.2$ & $82.5 \pm 1.0$ \\
& \textbf{Attention 16} & $\bm{73.6 \pm 2.1}$ & $95.8 \pm 0.1$ & $83.4 \pm 0.9$ \\
\hline
\end{tabular}
\end{table}

First, we report results for our novel patch-based attention architecture in comparison to other methods, see Table~\ref{tab:resatt}. The attention-based method improves the MC-sensitivity by up to $\SI{7}{\percent}$. The previous method of global context modeling with recurrent units is substantially outperformed by our approach. 

Next, we visualize the learned attention weights which represent the importance that is given to each patch, see Figure~\ref{fig:vis}. Patches with a brighter border receive a higher weighting. In general, patches with the lesion or part of the lesion have a higher weighting. This appears to be independent of the lesion's location in the full image.

\begin{figure}
	\centering
	\includegraphics[width=0.8\linewidth]{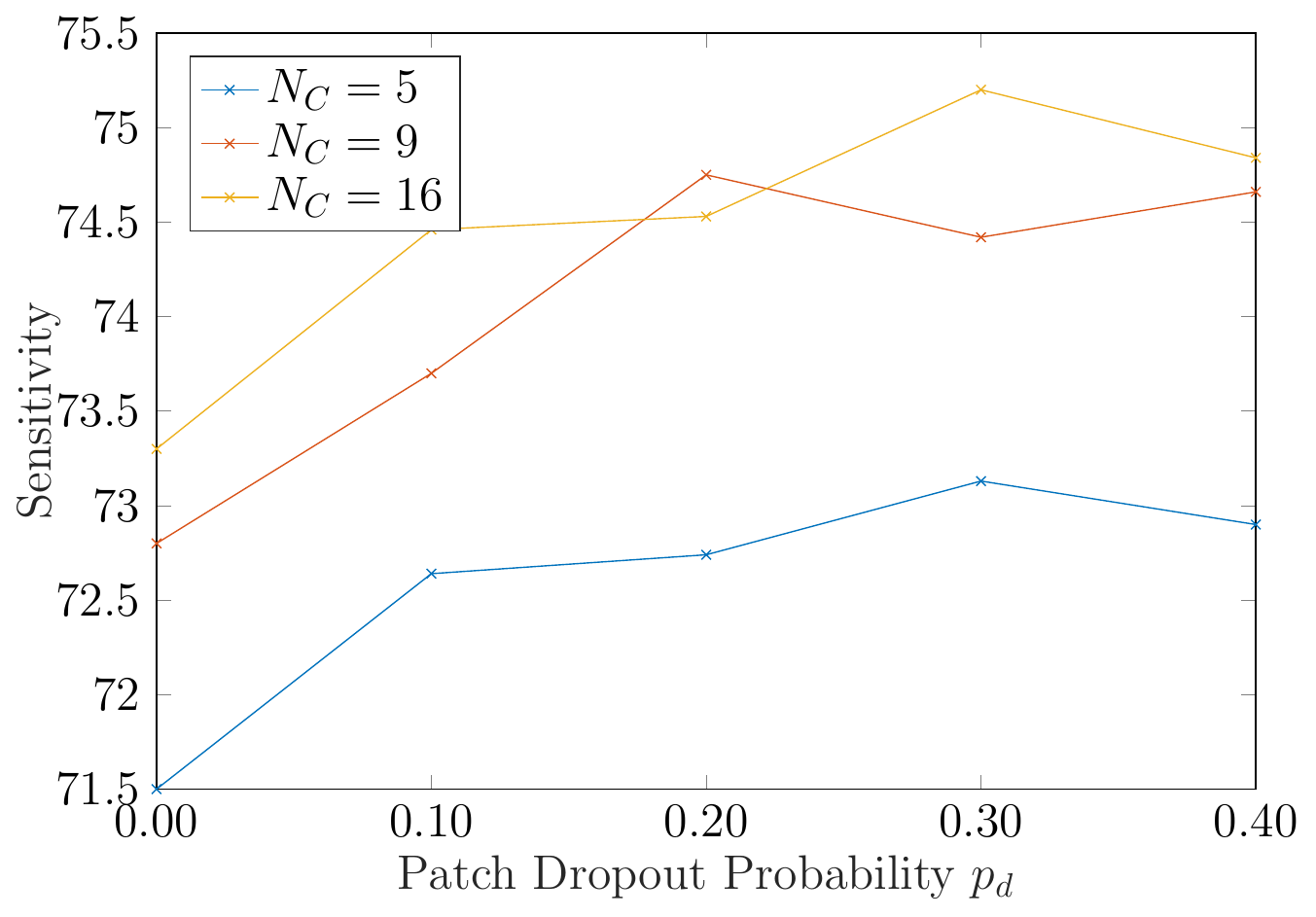}
   \caption{The impact of patch dropout on performance for Densenet121 with initial attention.}
\label{fig:patch_drop}	
\end{figure}

Furthermore, we consider varying numbers of crops. The results are shown in Table~\ref{tab:numcrops}. While adding more crops increases performance for Multi-Crop, moving from $9$ to $16$ crops leads to minor differences for the attention model. It should be noted that the attention model with $5$ crops achieves almost the same performance as the use of $16$ crops with Multi-Crop while only having to process a third of the number of crops.

Next, we perform evaluation on the SPC dataset. The results are shown in Table~\ref{tab:set}. Generally, the performance is slightly lower than for HAM. The relative performance differences between the methods are similar to HAM with the attention mechanism performing best.

\begin{table}[!t]
\renewcommand{\arraystretch}{1.3}
\caption{Results for evaluation on the test set of SPC. The attention model uses initial attention.}
\label{tab:set}
\centering
\begin{tabular}{l l l l l}
 & & MC-Sensitivity & MC-Specificity & F1-Score \\
\hline
\parbox[t]{2mm}{\multirow{3}{*}{\rotatebox[origin=c]{90}{InceptV3}}} & Multi-Crop & $59.7$ & $91.8$ & $71.1$ \\
& GRU & $62.0$ & $92.2$ & $72.4$ \\
& \textbf{Attention} & $\bm{64.0}$ & $93.1$ & $75.5$ \\
\hline
\parbox[t]{2mm}{\multirow{3}{*}{\rotatebox[origin=c]{90}{Dense}}} & Multi-Crop & $63.4$ & $93.0$ & $75.8$ \\
& GRU & $66.4$ & $92.5$ & $75.8$ \\
& \textbf{Attention} & $\bm{67.8}$ & $93.1$ & $75.2$ \\
\hline
\parbox[t]{2mm}{\multirow{3}{*}{\rotatebox[origin=c]{90}{SE-RX50}}} & Multi-Crop & $63.4$ & $92.3$ & $72.4$ \\
& GRU & $64.2$ & $92.2$ & $71.3$ \\
& \textbf{Attention} & $\bm{66.9}$ & $93.2$ & $73.0$ \\
\hline
\end{tabular}
\end{table}

Additionally, we show the effect of our novel patch dropout mechanism, see Fig.~\ref{fig:patch_drop}. For all $N_C$, patch dropout improves performance considerably. With increasing $p_d$, performance first improves and declines slightly at larger values.

\begin{table}[!t]
\renewcommand{\arraystretch}{1.3}
\caption{Results for different balancing strategies. Diag. weighting refers to our novel weighting strategy.}
\label{tab:bal}
\centering
\begin{tabular}{l l l l l}
 & & MC-Sensitivity & MC-Specificity & F1-Score \\
\hline
\parbox[t]{2mm}{\multirow{5}{*}{\rotatebox[origin=c]{90}{InceptV3}}} & No Balancing & $59.0 \pm 2.4$ & $94.5 \pm 0.3$ & $84.7 \pm 0.3$ \\
& Oversampling & $60.1 \pm 1.0$ & $94.9 \pm 0.2$ & $83.7 \pm 0.5$ \\
& Balanced Batches & $64.8 \pm 1.75$ & $95.5 \pm 0.1$ & $80.4 \pm 0.4$ \\
& Loss Weighting & $68.4 \pm 2.8$ & $95.5 \pm 0.2$ & $78.2 \pm 0.2$ \\
& \textbf{Diag. Weighting} & $\bm{70.0 \pm 1.1}$ & $95.4 \pm 0.2$ & $77.9 \pm 0.5$ \\
\hline
\parbox[t]{2mm}{\multirow{5}{*}{\rotatebox[origin=c]{90}{Dense121}}} & No Balancing & $65.0 \pm 1.0$ & $95.2 \pm 0.2$ & $85.5 \pm 0.6$ \\
& Oversampling & $65.7 \pm 1.7$ & $95.8 \pm 0.5$ & $84.5 \pm 0.2$ \\
& Balanced Batches & $68.4 \pm 0.6$ & $95.9 \pm 0.2$ & $82.4 \pm 0.4$ \\
& Loss Weighting & $71.1 \pm 1.1$ & $96.1 \pm 0.1$ & $82.1 \pm 0.3$ \\
& \textbf{Diag. Weighting} & $\bm{72.9 \pm 0.9}$ & $96.1 \pm 0.1$ & $80.2 \pm 0.8$ \\
\hline
\parbox[t]{2mm}{\multirow{5}{*}{\rotatebox[origin=c]{90}{SE-RX}}} & No Balancing & $63.7 \pm 0.4$ & $95.0 \pm 0.3$ & $85.4 \pm 0.3$ \\
& Oversampling & $61.3 \pm 2.4$ & $95.2 \pm 0.2$ & $84.0 \pm 0.5$ \\
& Balanced Batches & $68.9 \pm 0.9$ & $95.9 \pm 0.7$ & $82.3 \pm 1.2$ \\
& Loss Weighting & $70.2 \pm 1.0$ & $96.0 \pm 0.0$ & $81.2 \pm 0.7$ \\
& \textbf{Diag. Weighting} & $\bm{72.4 \pm 1.0}$ & $95.9 \pm 0.1$ & $78.6 \pm 1.2$ \\
\hline
\end{tabular}
\end{table}

Next, we address various class balancing techniques for the problem at hand, see Table~\ref{tab:bal}. The strict balanced sampling shows a clear performance improvement while simple oversampling yields similar or worse results. Loss weighting results in slightly better performance than balanced sampling and our novel diagnosis-guided loss weighting strategy leads to the overall best performance.

\begin{table}[!t]
\renewcommand{\arraystretch}{1.3}
\caption{Results for different values of $k$ for loss weighting.}
\label{tab:weighting}
\centering
\begin{tabular}{l l l l l}
 & & MC-Sensitivity & MC-Specificity & F1-Score \\
\hline
\parbox[t]{2mm}{\multirow{5}{*}{\rotatebox[origin=c]{90}{IncepV3}}} & $k=0.50$ & $65.0 \pm 1.5$ & $95.3 \pm 0.2$ & $83.9 \pm 1.0$ \\
& $k=0.75$ & $68.4 \pm 2.1$ & $95.6 \pm 0.3$ & $81.5 \pm 0.1$ \\
& $k=1.00$ & $68.4 \pm 2.8$ & $95.5 \pm 0.2$ & $78.2 \pm 0.2$ \\
& $\bm{k=1.25}$ & $\bm{69.3 \pm 0.4}$ & $95.1 \pm 0.1$ & $71.1 \pm 1.6$ \\
& $k=1.50$ & $68.9 \pm 1.7$ & $93.8 \pm 0.1$ & $58.8 \pm 1.3$ \\
\hline
\parbox[t]{2mm}{\multirow{5}{*}{\rotatebox[origin=c]{90}{Dense121}}} & $k=0.50$ & $68.7 \pm 0.6$ & $95.9 \pm 0.1$ & $85.4 \pm 0.3$ \\
& $k=0.75$ & $70.1 \pm 0.5$ & $96.0 \pm 0.1$ & $84.3 \pm 0.3$ \\
& $k=1.00$ & $71.1 \pm 1.1$ & $96.1 \pm 0.1$ & $82.1 \pm 0.3$ \\
& $\bm{k=1.25}$ & $\bm{72.2 \pm 1.1}$ & $95.9 \pm 0.1$ & $78.6 \pm 1.9$ \\
& $k=1.50$ & $71.8 \pm 0.4$ & $95.4 \pm 0.1$ & $71.2 \pm 1.6$ \\
\hline
\parbox[t]{2mm}{\multirow{5}{*}{\rotatebox[origin=c]{90}{SE-RX50}}} & $k=0.50$ & $68.0 \pm 2.2$ & $95.7 \pm 0.3$ & $84.5 \pm 0.6$ \\
& $k=0.75$ & $69.7 \pm 2.0$ & $95.9 \pm 0.6$ & $83.7 \pm 0.2$ \\
& $k=1.00$ & $70.2 \pm 1.0$ & $96.0 \pm 0.0$ & $81.2 \pm 0.7$ \\
& $k=1.25$ & $70.4 \pm 1.2$ & $95.4 \pm 0.4$ & $74.8 \pm 0.6$ \\
& $\bm{k=1.50}$ & $\bm{70.8 \pm 0.9}$ & $94.7 \pm 0.2$ & $65.5 \pm 1.7$ \\
\hline
\end{tabular}
\end{table}

Last, we examine varying strengths of loss weighting by varying the exponent $k$, see Table~\ref{tab:weighting}. Starting with $k<1$, performance first increases with increasing $k$ but the MC-sensitivity quickly saturates for values $k>1$. Note that while the MC-sensitivity improves or stays the same with large $k$, the F1-score starts to deteriorate. 
Overall, the MC-specificity remains at a high level across all experiments with only minor variations. 

\section{Discussion} \label{sec:discussion}

In the past, skin lesion diagnosis has largely been focused on the binary decision problem of whether or not to perform a biopsy. Recently, more fine-grade distinction of lesions from dermoscopy images has been brought up as a challenge with the publicly available HAM dataset \cite{tschandl2018}. The more fine-grained differentiation of lesions is closer to a dermatologists workflow \cite{nachbar1994abcd,argenziano1998epiluminescence} which, however, comes with new challenges for skin lesion classification. For adequate assessment of model performance for the multi-class problem, we consider the MC-sensitivity as a key metric. In contrast to other metrics, the MC-sensitivity is independent of the number of examples per class and it provides insight on how well each lesion can be differentiated.

\textbf{Using high-resolution images} is challenging with deep learning methods as memory and computational resources are still limited. To find more fine-grained differences between lesions, high-resolution image information is likely helpful and should be made use of when designing high-performance models. A simple approach is to use multi-crop evaluation, i.e. a number of small, high-resolution crops is fed into standard models and the crops' predictions are averaged. In this paper, we extend this approach by also incorporating global context between the local patches with a patch-based attention mechanism. The attention block can be plugged into any pretrained architecture and thus allows for easy exploitation of transfer learning performance as well as the advantage of using high-resolution images. Our results show that the proposed module improves performance substantially, see Table~\ref{tab:resatt}. In particular, the previous approach of global context modeling with recurrent blocks is outperformed. Figure~\ref{fig:vis} provides a qualitative view on the learned attention weights. Higher weighting of patches which contain the lesion indicate that the model learned to focus on the relevant patches. It is notable that the model does not focus entirely on a single patch where most of the lesion is visible but also puts weight on patches with the lesion border and smaller parts of the lesion. This relates to typical diagnostic criteria where, e.g., streaks at the lesion border are an important feature \cite{argenziano1998epiluminescence}. These results indicate that our attention mechanism enables effective global information exchange and focus on the relevant parts of the image.

The effect of global context recovery becomes a lot more pronounced when considering fewer crops, that do not cover the entire image, see Table~\ref{tab:numcrops}. For five crops, the attention mechanism's MC-sensitivity is only $0.2-0.5$ percent points lower than averaging with $16$ crops. Thus, the attention model can make predictions three times as fast with similar accuracy which is a key aspect considering the large amount of lesions that need to be diagnosed in clinical practice. Also, it is notable that the performance increase between attention models with $9$ and $16$ crops is very small. Considering that $9$ crops already cover the entire image, this underlines effective exploitation of the available, global information and adding more (partially redundant) patches is not required. Overall, our novel patch-based attention mechanism improves the MC-sensitivity by $\approx \SI{7}{\percent}$ while only increasing the number of parameters by $\approx \SI{0.001}{\percent}$. 

To further improve performance, we also propose \textit{patch dropout} which appears to be an effective regularization technique, see Fig.~\ref{fig:patch_drop}. The fixed order of crops for patch-based methods could lead to overfitting to particular patch positions. Our patch dropout method prevents this problem by forcing the models to deal with occasionally missing patches.

Our attention mechanism learns the importance of patches and might therefore be biased to the particular zoom and manual cropping performed by the curators of HAM. Therefore, we also apply the method to the SPC dataset \cite{kawahara20187}, see Table~\ref{tab:set}. The overall performance is slightly lower than for HAM, likely due to the smaller training set size. Also, some of the images contain artifacts such as a reference grid which might also reduce performance. Still, the relative performance improvement of our attention mechanism over other patch combination techniques remains.
It should be noted that the improvement mostly shows up in the MC-sensitivity and not as much for the MC-specificity and F1-score.

\textbf{Dealing with extreme class imbalance} is another problem that arises from more fine-grained multi-class classification with a highly unbalanced real-world skin lesion dataset such as HAM. So far, dealing with class imbalance has been considered a minor problem that is hardly addressed for skin lesion classification \cite{kawahara20187}. However, the imbalance is crucial when considering the MC-sensitivity of a model. This is indicated in Table~\ref{tab:bal} where using no measure to counter class imbalance leads to a poor MC-sensitivity. Notably, this effect does not occur in the other metrics, e.g., the F1-score remains high. This underlines the importance of the MC-sensitivity as a key metric to identify model performance with respect to multi-class lesion diagnosis. 
Regarding sampling strategies, strict sampling appears to be advantageous over simple oversampling, see Table~\ref{tab:bal}. While oversampling does not lead to a consistent performance improvement, balanced sampling leads to an improvement of $\SI{5}{\percent}$ to $\SI{8}{\percent}$ for the MC-sensitivity. This is notable in particular, as classes appear in an overall equal amount and individual, balanced updates appear to be important to reach a good local minimum. 
Using a loss weighting strategy improves the MC-sensitivity by $\SI{10}{\percent}$ over no balancing at all and by $\SI{2}{\percent}$ to $\SI{4}{\percent}$ over balanced sampling. When applying our novel diagnosis-guided weighting on top, performance improves further by $\approx\SI{3}{\percent}$. Although the performance gain is comparatively small, this indicates that the method of diagnosis does contain relevant information for the learning problem and in the future more effective ways to exploit it could be developed.
When considering the results for different strengths of loss weighting in Table~\ref{tab:weighting}, it is also evident that a plain increase in the weighting exponent $k$ cannot achieve the same effect as the diagnosis-guided weighting. Overall, stronger balancing improves the MC-sensitivity, however, the F1-score deteriorates considerably with increasing $k$ which indicates that a high MC-sensitivity and thus better class distinction needs to be traded off for other performance metrics. In this regard, it is notable that for the attention-based models the increase in MC-sensitivity is not bought by a decrease in other metrics. 
Regarding the MC-specificity of the different model variants, we find little variance. The metric shows consistently high values which underlines that the MC-sensitivity is the key problem to consider.

Overall, we performed our experiments for multiple and different state-of-the-art architectures. Key insights are supported by consistent results across all models.

\section{Conclusion} \label{sec:conclusion}

We address two key properties of multi-class skin lesion classification based on dermoscopic image data. First, we make use of the high-resolution images with state-of-the-art pretrained CNN architectures and different approaches to combine local and global context. We show that a novel patch-based attention method to handle multiple overlapping crops yields substantially better performance for three network architectures and two datasets. The second problem is the severe class imbalance in multi-class skin lesion datasets. We compare different approaches for sampling and weighting schemes during training. We show that a new method using the type of ground-truth to weight the loss function results in improved classification performance. Both problems, high resolution image data and class imbalance, are frequently encountered in medical diagnosis and hence future work could study the proposed methods in the context of other clinical applications.

\section*{Acknowledgment}

The authors would like to thank NVIDIA for the donation of a graphics card under the GPU Grant Program. This work was partially supported by the Forschungszentrum Medizintechnik Hamburg (02fmthh2017).
\ifCLASSOPTIONcaptionsoff
  \newpage
\fi



%\bibliographystyle{IEEEtran}
%\begin{thebibliography}{1}


%\end{thebibliography}
\bibliographystyle{IEEEtran} 
\bibliography{egbib}

\begin{thebibliography}{10}
\providecommand{\url}[1]{#1}
\csname url@samestyle\endcsname
\providecommand{\newblock}{\relax}
\providecommand{\bibinfo}[2]{#2}
\providecommand{\BIBentrySTDinterwordspacing}{\spaceskip=0pt\relax}
\providecommand{\BIBentryALTinterwordstretchfactor}{4}
\providecommand{\BIBentryALTinterwordspacing}{\spaceskip=\fontdimen2\font plus
\BIBentryALTinterwordstretchfactor\fontdimen3\font minus
  \fontdimen4\font\relax}
\providecommand{\BIBforeignlanguage}[2]{{%
\expandafter\ifx\csname l@#1\endcsname\relax
\typeout{** WARNING: IEEEtran.bst: No hyphenation pattern has been}%
\typeout{** loaded for the language `#1'. Using the pattern for}%
\typeout{** the default language instead.}%
\else
\language=\csname l@#1\endcsname
\fi
#2}}
\providecommand{\BIBdecl}{\relax}
\BIBdecl

\bibitem{siegel2017cancer}
R.~L. Siegel~et al., ``Cancer statistics, 2017,'' \emph{CA: a cancer journal
  for clinicians}, vol.~67, no.~1, pp. 7--30, 2017.

\bibitem{apalla2017skin}
Z.~Apalla~et al., ``Skin cancer: Epidemiology, disease burden, pathophysiology,
  diagnosis, and therapeutic approaches,'' \emph{Dermatology and therapy},
  vol.~7, no.~1, pp. 5--19, 2017.

\bibitem{kittler2002diagnostic}
H.~Kittler~et al., ``Diagnostic accuracy of dermoscopy,'' \emph{The lancet
  oncology}, vol.~3, no.~3, pp. 159--165, 2002.

\bibitem{vestergaard2008dermoscopy}
M.~Vestergaard~et al., ``Dermoscopy compared with naked eye examination for the
  diagnosis of primary melanoma: a meta-analysis of studies performed in a
  clinical setting,'' \emph{British Journal of Dermatology}, vol. 159, no.~3,
  pp. 669--676, 2008.

\bibitem{argenziano1998epiluminescence}
G.~Argenziano~et al., ``Epiluminescence microscopy for the diagnosis of
  doubtful melanocytic skin lesions: comparison of the abcd rule of
  dermatoscopy and a new 7-point checklist based on pattern analysis,''
  \emph{Archives of Dermatology}, vol. 134, no.~12, pp. 1563--1570, 1998.

\bibitem{nachbar1994abcd}
F.~Nachbar~et al., ``{The ABCD rule of dermatoscopy: high prospective value in
  the diagnosis of doubtful melanocytic skin lesions},'' \emph{Journal of the
  American Academy of Dermatology}, vol.~30, no.~4, pp. 551--559, 1994.

\bibitem{kimball2008us}
A.~B. Kimball and J.~S. Resneck~Jr, ``The us dermatology workforce: a specialty
  remains in shortage,'' \emph{Journal of the American Academy of Dermatology},
  vol.~59, no.~5, pp. 741--745, 2008.

\bibitem{maragoudakis2010skin}
M.~Maragoudakis and I.~Maglogiannis, ``Skin lesion diagnosis from images using
  novel ensemble classification techniques,'' in \emph{Information Technology
  and Applications in Biomedicine (ITAB), 2010 10th IEEE International
  Conference on}.\hskip 1em plus 0.5em minus 0.4em\relax IEEE, 2010, pp. 1--5.

\bibitem{madooei2012intrinsic}
A.~Madooei~et al., ``Intrinsic melanin and hemoglobin colour components for
  skin lesion malignancy detection,'' in \emph{MICCAI}.\hskip 1em plus 0.5em
  minus 0.4em\relax Springer, 2012, pp. 315--322.

\bibitem{litjens2017survey}
G.~Litjens~et al., ``A survey on deep learning in medical image analysis,''
  \emph{Medical Image Analysis}, vol.~42, pp. 60--88, 2017.

\bibitem{codella2015deep}
N.~Codella~et al., ``Deep learning, sparse coding, and svm for melanoma
  recognition in dermoscopy images,'' in \emph{International Workshop on
  Machine Learning in Medical Imaging}.\hskip 1em plus 0.5em minus 0.4em\relax
  Springer, 2015, pp. 118--126.

\bibitem{kawahara2016deep}
J.~Kawahara~et al., ``Deep features to classify skin lesions,'' in \emph{ISBI},
  2016, pp. 1397--1400.

\bibitem{kawahara2016multi}
J.~Kawahara and G.~Hamarneh, ``{Multi-resolution-tract CNN with hybrid
  pretrained and skin-lesion trained layers},'' in \emph{International Workshop
  on Machine Learning in Medical Imaging}.\hskip 1em plus 0.5em minus
  0.4em\relax Springer, 2016, pp. 164--171.

\bibitem{lopez2017skin}
A.~R. Lopez~et al., ``Skin lesion classification from dermoscopic images using
  deep learning techniques,'' in \emph{Biomedical Engineering (BioMed), 2017
  13th IASTED International Conference on}.\hskip 1em plus 0.5em minus
  0.4em\relax IEEE, 2017, pp. 49--54.

\bibitem{yang2018clinical}
J.~Yang~et al., ``Clinical skin lesion diagnosis using representations inspired
  by dermatologist criteria,'' in \emph{CVPR}, vol.~11, 2018.

\bibitem{kawahara20187}
J.~Kawahara~et al., ``7-point checklist and skin lesion classification using
  multi-task multi-modal neural nets,'' \emph{IEEE Journal of Biomedical and
  Health Informatics}, 2018.

\bibitem{diaz2018dermaknet}
I.~G. Diaz, ``Dermaknet: Incorporating the knowledge of dermatologists to
  convolutional neural networks for skin lesion diagnosis,'' \emph{IEEE Journal
  of Biomedical and Health Informatics}, 2018.

\bibitem{esteva2017dermatologist}
A.~Esteva~et al., ``Dermatologist-level classification of skin cancer with deep
  neural networks,'' \emph{Nature}, vol. 542, no. 7639, p. 115, 2017.

\bibitem{Szegedy.2016b}
C.~Szegedy~et al., ``{Rethinking the inception architecture for computer
  vision},'' in \emph{{CVPR}}, 2016, pp. 2818--2826.

\bibitem{isicdatabase}
``{The International Skin Imaging Collaboration: Melanoma Project},''
  \url{https://www.isic-archive.com/}, accessed: 2018-09-13.

\bibitem{codella2018skin}
N.~C. Codella~et al., ``Skin lesion analysis toward melanoma detection: A
  challenge at the 2017 international symposium on biomedical imaging (isbi),
  hosted by the international skin imaging collaboration (isic),'' in
  \emph{ISBI}, 2018, pp. 168--172.

\bibitem{Argenziano}
G.~Argenziano~et al., \emph{Interactive atlas of dermoscopy. Dermoscopy: a
  tutorial (Book) and CD-ROM.}, 02 2000.

\bibitem{tschandl2018}
P.~Tschandl~et al., ``{The HAM10000 dataset, a large collection of multi-source
  dermatoscopic images of common pigmented skin lesions},'' \emph{Scientific
  Data}, vol.~5, no. 180161, 2018.

\bibitem{gessert2018skin}
N.~Gessert~et al., ``Skin lesion diagnosis using ensembles, unscaled multi-crop
  evaluation and loss weighting,'' \emph{arXiv preprint arXiv:1808.01694},
  2018.

\bibitem{Kamnitsas.2017}
K.~Kamnitsas~et al., ``{Efficient multi-scale 3D CNN with fully connected CRF
  for accurate brain lesion segmentation},'' \emph{{Medical Image Analysis}},
  vol.~36, pp. 61--78, 2017.

\bibitem{hoo2016deep}
S.~Hoo-Chang~et al., ``Deep convolutional neural networks for computer-aided
  detection: Cnn architectures, dataset characteristics and transfer
  learning,'' \emph{IEEE Transactions on Medical Imaging}, vol.~35, no.~5, p.
  1285, 2016.

\bibitem{tajbakhsh2016convolutional}
N.~Tajbakhsh, J.~Y. Shin, S.~R. Gurudu, R.~T. Hurst, C.~B. Kendall, M.~B.
  Gotway, and J.~Liang, ``Convolutional neural networks for medical image
  analysis: Full training or fine tuning?'' \emph{IEEE Transactions on Medical
  Imaging}, vol.~35, no.~5, pp. 1299--1312, 2016.

\bibitem{zhu2017dependency}
X.~Zhu~et al., ``{Dependency exploitation: a unified CNN-RNN approach for
  visual emotion recognition},'' in \emph{Proceedings of the Internal Joint
  Conference on Artificial Intelligence (IJCAI 2017)}, 2017.

\bibitem{li2016lstm}
Z.~Li~et al., ``{LSTM-CF: Unifying context modeling and fusion with lstms for
  RGB-D scene labeling},'' in \emph{ECCV}.\hskip 1em plus 0.5em minus
  0.4em\relax Springer, 2016, pp. 541--557.

\bibitem{hu2018squeeze}
J.~Hu~et al., ``Squeeze-and-excitation networks,'' in \emph{CVPR}, 2018.

\bibitem{Litjens.2017}
G.~Litjens, T.~Kooi, B.~E. Bejnordi, A.~A.~A. Setio, F.~Ciompi, M.~Ghafoorian,
  J.~A. {van der Laak}, B.~{van Ginneken}, and C.~I. S{\'a}nchez, ``{A survey
  on deep learning in medical image analysis},'' \emph{{arXiv preprint
  arXiv:1702.05747}}, 2017.

\bibitem{huang2017densely}
G.~Huang~et al., ``Densely connected convolutional networks,'' in \emph{CVPR},
  2017.

\bibitem{Ioffe.2015}
S.~Ioffe and C.~Szegedy, ``Batch normalization: Accelerating deep network
  training by reducing internal covariate shift,'' in \emph{ICML}, 2015, pp.
  448--456.

\bibitem{Xie.2017}
S.~Xie~et al., ``Aggregated residual transformations for deep neural
  networks,'' in \emph{CVPR}, 2017, pp. 5987--5995.

\bibitem{cho2014learning}
K.~Cho~et al., ``Learning phrase representations using rnn encoder-decoder for
  statistical machine translation,'' in \emph{EMNLP}, 2014.

\bibitem{Kingma.2014}
D.~Kingma and J.~Ba, ``{Adam: A method for stochastic optimization},'' in
  \emph{{ICLR}}, 2014.

\bibitem{paszke2017automatic}
A.~Paszke~et al., ``Automatic differentiation in pytorch,'' in \emph{NIPS-W},
  2017.

\end{thebibliography}

%\end{IEEEbiography}


%\newpage



%\vfill


\end{document}